# Deep learning-based approach to reveal tumor mutational burden status from whole slide images across multiple cancer types

Siteng Chen[1*], Jinxi Xiang[2*], Xiyue Wang[3*], Jun Zhang[2#], Sen Yang[2], Junzhou Huang[2], Wei Yang[2], Junhua Zheng[1#], Xiao Han[2#]

[1] Department of Urology, Renji Hospital, Shanghai Jiao Tong University School of Medicine, Shanghai 200135, China.
[2] Tencent AI Lab, Shenzhen 518057, China.
[3] College of Computer Science, Sichuan University, Chengdu 610065, China.

*Equal contributors and co-first authors

#Corresponding authors:
Jun Zhang, Tencent AI Lab, Shenzhen 518057, China. E-mail: junejzhang@tencent.com. Tel:86-0755-86013388-887069.
Junhua Zheng, Department of Urology, Renji Hospital, Shanghai Jiao Tong University School of Medicine, Shanghai 200080, China. E-mail: Zhengjh1900@163.com. Tel: 86-021-63240090.
Xiao Han, Tencent AI Lab, Shenzhen 518057, China. E-mail: haroldhann@tencent.com. Tel: 86-0755-86013388-779087.

**Abstract**

Tumor mutational burden (TMB) is a potential genomic biomarker of immunotherapy. However, TMB detected through whole exome sequencing lacks clinical penetration in low-resource settings. In this study, we proposed a multi-scale deep learning framework to address the detection of TMB status from routinely used whole slide images for a multiple cancer TMB prediction model (MC-TMB). The MC-TMB achieved a mean area under the curve (AUC) of 0.818 (0.804-0.831) in the cross-validation cohort, which showed superior performance to each single-scale model. The improvements of MC-TMB over the single-tumor models were also confirmed by the ablation tests on ×10 magnification, and the highly concerned regions typically correspond to dense lymphocytic infiltration and heteromorphic tumor cells. MC-TMB algorithm also exhibited good generalization on the external validation cohort with an AUC of 0.732 (0.683-0.761), and better performance when compared to other methods. In conclusion we proposed a deep learning-based approach to reveal tumor mutational burden status from routinely used pathological slides across multiple cancer types.

**INTRODUCTION**

The development of immunotherapy has been reported as an important breakthrough in the treatment of cancer. The expression of programmed cell death 1 ligand in tumor cells,[1] microsatellite instability, and tumor mutational burden (TMB) [2, 3] have been currently reported to act as effective biomarkers for immune-checkpoint inhibitor (ICI) therapy. However, only TMB has been proved to serve as a candidate biomarker of clinical outcomes from immunotherapy for multiple solid tumors [4, 5]. A higher level of TMB is associated with increased new antigens, which could be easier recognized by the immune system and help benefit from immunotherapy [6].

TMB is calculated by counting the number of nonsynonymous mutations across a length of genome sequenced through whole-exome sequencing (WES), which is reported as mutations per coding area of a tumor genome [7]. Despite the outstanding performance in predicting immunotherapeutic responsiveness, TMB is not widely available in clinical applications due to the high cost, operational complexity, and lengthy time for WES [8]. Even though TMB can also be evaluated from lower-cost panels with targeted genes, potential bias could exist due to the limited fraction of the exome sequenced [7]. Therefore, a scalable and time-efficient approach for the identification of TMB status with low cost is desirable for patients.

Computational histopathology algorithms, often based on convolutional neural networks (CNN), can process and cross-reference large volumes of data to aid in quantifying aberrant cells and tissues [9]. Recent advances in deep learning from hematoxylin-eosin (H&E) stained whole slide images (WSI) have also demonstrated satisfactory performance in capturing phenotypes that are typically not recognized by experienced pathologists [10, 11]. Establishing associations of a digitized WSI with high or low TMB without detailed annotations at the cellular and regional levels is typically considered a weakly-supervised learning problem, and the challenge lies in determining visual clues in slides associated with specific gene status. Currently, related works for TMB prediction from H&E-stained slides only focused on single cancer, such as stomach adenocarcinoma, colon adenocarcinoma [12], or lung adenocarcinoma [13], separately. However, current studies omitted the quantified associations with traits beyond cancer types, failing to provide the required generalization performance across various tumor types. The investigation of the multiple cancer model can potentially reduce the work-ups to develop customed models for each cancer type, and could also probably enhance the overall prediction performance by exploiting larger datasets and universal features.

In this study, we aimed to propose a novel weakly-supervised deep learning method with spatial awareness for predicting TMB of a range of tumors based only on H&E-stained sections and patient-level labels. To our knowledge, this is the first study to propose a state-of-the-art framework for TMB prediction from histopathological images of multiple solid tumors, which might promote the potential utility of these neural networks in exploring biomarkers for immunotherapy.

## RESULTS

### Multiple cancer tumor mutational burden assessment via deep learning

The multiple cancer TMB prediction model (MC-TMB) was developed on an internal dataset consisting of 3228 gigapixel WSIs from the Cancer Genome Atlas (TCGA) cohort. Each slide had an associated label for identifying the status of TMB as high or low, which was confirmed through WES. MC-TMB updated the feature vector by aggregating feature information from its neighbors in the graph based on a multi-scale approach across three magnifications (×5, ×10, and ×20). Five-fold cross-validation was performed for the training and internal validation of the model at the slide level to reduce overfitting and improve the prediction stability. To rigorously assess model adaptability, MC-TMB was tested on an additional independent dataset from the Clinical Proteomic Tumor Analysis Consortium (CPTAC) cohort (531 WSIs from 5 tumor types).

### Overall performance of the MC-TMB

MC-TMB achieved a mean area under the curve (AUC) of 0.818 (0.804-0.831) in the five-fold cross-validation of the TCGA cohort, with a sensitivity of 0.715 (0.684-0.745) and specificity of 0.756 (0.738-0.773, Fig. 2A). The overall performance showed the clinical significance of TMB predictions from the MC-TMB, and two highly accurately predicted TMB were colon adenocarcinoma (COAD) and stomach adenocarcinoma (STAD), with AUCs of 0.899 and 0.878, respectively. Assessment of model performance on patient subgroups (Table S1) further showed the model's robustness to variations in patient age, sex, tumor grade, and tumor stage.

An ablation study was further conducted on multi-scale training against single-scale, arguing that multiple cancer training with multi-scale could greatly improve the model performance. We examined model performance across different magnifications. The prediction model achieved an AUC of 0.721 (0.705-0.737) on ×5 magnification scale, 0.771 (0.756-0.785) on ×10 magnification, and 0.787 (0.772-0.801) on ×20 magnification. Significant improvement of the multi-scale model (MC-TMB) was observed compared to each single-scale model (Fig. 3, p < 0.001), suggesting that there was complementary information that the deep learning algorithm could capture from different levels of granularity in the images. Further subgroup analysis also illustrated the superiority of multi-scale training in each tumor type, indicating that the MC-TMB algorithm might capture more favorable features from different levels of granularity in the WSIs.

### Model interpretability

To deliver more interpretable insights into MC-TMB, we generated attention heatmaps by converting the attention scores to their corresponding spatial locations in the whole slide. The attention score, which represented the interpretation of the predicted relevance of each slide region, was translated into patch-level visualization to create an attention heatmap and displayed as a semitransparent overlay (Fig. 4). It could be hard to interpret these heatmaps across multiple cancer types manual annotations, but the model had learned to distinguish some shared features. For MC-TMB, the highly concerned regions typically correspond to dense lymphocytic infiltration and heteromorphic tumor cells, whereas the low attention scores corresponded to regions with normal tissue or dispersive tumor cells without lymphocytic infiltration. In addition, the patch-level visualization of MC-TMB also demonstrated spatial heterogeneity of TMB within tumor samples.

**Multiple cancer task outperforms single-cancer task**

One can choose to train an independent model for each cancer type as most previous studies did. Instead, we trained a unified model using data from all cancer types simultaneously, trying to discover novel associations of visual traits across different cancer. To this end, another ablation study was performed to argue the improvements in identifying TMB status by MC-TMB over the single-cancer models. For comparison, we trained 7 individual deep-learning models for each cancer type using the same weakly-supervised network on ×10 magnification.

Remarkably, we found that the multiple cancer model outperformed single cancer model in predicting TMB status (AUC: 0.771 vs. 0.714, p < 0.001, Fig. 5A). Subgroup analysis based on tumor types also illustrated a superior predictive accuracy of the multiple cancer model compared with respective single-cancer model (Fig. 5B). To further investigate the advantage of the multiple cancer model, we carried out a detailed analysis of failure cases of a single-cancer model, but not multiple cancer model. The representative misclassified cases from the single-cancer model were detailed in Fig. 5C. We observed that for these failure cases, the single-cancer model might fail to recognize the regions with nontypical lymphocytic infiltration. However, the multiple cancer model could still assign high attention scores to the regions with nontypical lymphocytic infiltration and made correct predictions. One such possibility could be a reliance upon the inclusion of class labels to the classifier. This observation was consistent with the standpoint, where computational feature representation could find similar images and quantify associations with traits beyond tissue types [14].

**Validation of external cohort**

The multi-scale MC-TMB model was applied to the independent international CPTAC cohort without any form of domain adaptation or model tuning. The results indicated that MC-TMB also exhibited good generalization in the external cohort with an AUC of 0.732 (0.683-0.761, Fig. 2B). The prediction of TMB status was well replicated in the vast majority of cancer types in the external cohort, especially for lung adenocarcinoma (LUAD) [AUC=0.778 (0.687-0.853)], COAD [AUC=0.719 (0.635-0.793)], and head and neck squamous cell carcinoma (HNSC) [AUC=0.705 (0.601-0.796)]. A moderate drop in accuracy was found in the external validation, which might result from the potential batch effect and the limited sample size of the validation cohort. In brief, these results further illustrated the demonstrates the ability of MC-TMB to generalize across diverse populations and different tumor types. The comparison of MC-TMB with other methods in the independent patient cohort was shown in Table 1. The results indicated that our model performed better in predicting TMB status from WSIs.

**Tumor immune microenvironment associated with MC-TMB**

Except for TMB, microsatellite instability (MSI) and mutations in specific genes were also associated with immunotherapy effect, such as DNA damage and repair (DDR) [15], TP53 [16], KRAS [17], and EGFR [18], which had also been reported to be highly correlated to tumor immunotherapy and deep learning-based image features [11]. We next explore the potential relationship between tumor immune microenvironment and MC-TMB. The threshold of MC-TMB (0.3205) was identified using the Youden index. Patients with a probability more than 0.3205 were regarded as MC-TMB (+), otherwise MC-TMB (-). As shown in Fig. 6A and Fig. S1A, a distinct tumor mutation spectrum could be observed between MC-TMB (+) and MC-TMB (-). MC-TMB (+) was associated with a higher mutation frequency of some immunotherapy-related genes, including TP53, KRAS, POLE, PBRM1, SK11, JAK1/2/3, POLD1, and IFNGR1 (Fig. 6B). Surprisingly, MC-TMB also performed well in distinguishing MSI-H from routine H&E-stained slides (Fig. 6C), with AUC of 0.767 (0.752-0.781), which might due to the high correlation between TMB and MSI-sensor score (Fig. S1B).

Subgroup analysis revealed that MC-TMB performed well in predicting the status of MSI and exhibited the model robustness to variations in tumor types, especially for HNSC, COAD, and STAD (Table S2). A higher percentage of C2 immunophenotyping was also found in the MC-TMB (+) group (Fig. 6D-E, Fig. S2), which was IFN-$\gamma$ dominant and overrode an evolving type I immune response [19].

Higher TMB and MSI-sensor scores were observed in the MC-TMB (+) group (Fig. 7A). Comparison of the infiltration level in each immune cell was illustrated in Fig. 7B. MC-TMB (+) was significantly associated with a higher infiltration level of CD8$^+$ T cell, which had been proved to be associated with the response to cancer immunotherapy [20]. In addition, higher levels of lymphocyte infiltration signature, PD-1 expression and PD-L1 expression were also observed in the MC-TMB (+) group (Fig. 7C). The different expressions of chemokines were also consistent with the results reported above (Fig. 7D). MC-TMB (+) was associated with higher expression of chemokines, including CXCL9, CXCL10 and CXCL13, which have been proved to attract CD8$^+$ T cells [21]. As shown in Fig. 7E, patients with different predicted TMB statuses had distinct prognosis during the follow-up of over ten years, with HR of 0.761(0.674-0.859). Meanwhile, immune-related hallmark pathways, such as regulation of immune response, were significantly enriched in the MC-TMB (+) group (Fig. 7F). All these results further expounded the potential guiding function of MC-TMB for tumor immunotherapy.

**DISCUSSIONS**

The predictive value of TMB for response to ICI therapy has been recognized in multiple cancers [2, 22, 23]. However, how to appropriately identifying TMB is still challenging since there are more than 30000 genes in a single cell and the calculation of various mutations in malignancies is intricate. Using the WES technology, a complete landscape of coding mutations could be acquired for the calculation of TMB. Nevertheless, the WES needs demanding samples and is strongly suggested for identifying tumor-specific variants from a tumor sample and matched normal samples, which incurs additional costs and time delays [24]. In addition, larger sizes of selected genes in some locally-accessible panels are still wanted for the normalized TMB evaluation [25].

The WES technique used to calculate TMB requires highly demanding samples, which results in additional cost and time delays. In this study, we proposed a novel MC-TMB based on weakly-supervised deep learning strategy for predicting TMB across multiple tumor types from H&E-stained slides. Taking about 4 minutes to scan the H&E-Stained slide and automatically analyze the WSI through our deep learning pipeline, the pathologist can evaluate the TMB status as soon as he or she makes the pathologic diagnosis.

Unlike previous studies only focusing on specific single tumor types [12, 13, 26, 27], our MC-TMB could be applied to 7 different types of cancers with satisfying prediction accuracy, especially in COAD and STAD. To our knowledge, this is the first study reporting a TMB prediction model across multiple solid tumors. MC-TMB achieved superior performance over previously reported models by a large margin on reported AUCs [12, 26, 27] (COAD: 0.899 vs 0.820; LUAD: 0.804 vs 0.710; STAD: 0.878 vs 0.750; bladder urothelial carcinoma (BLCA): 0.757 vs 0.750, with improvements of AUC from 0.5% to 12.8%. Computational feature representation could find similar images and quantify associations with traits beyond tissue types [14]. Therefore, better implementation in predicting TMB status might require generalization performance across various tumor types. Our study displayed the feasibility of multiple cancer learning-based inference of TMB status across multiple solid tumors directly from histological images, which provided perspectives evidences for the generalization of multi-carcinomatous species in genetic prediction.

Predicting the TMB status from the visual features of H&E slides without other clinical information is one kind of weakly supervised learning problem. To further verify the improvements of our model, we investigate the performance of the following readily available baselines using the same training/testing data. And we follow the original designs of these baselines to use ImageNet pretrained ResNet50 for preprocessing to conduct feature embeddings. AbMIL is the first attention-based model using the learnable network to calculate the weighted sum of all image patches for slide-level prediction. CLAM further improves AbMIL with clustering loss to affiliate the training of the attention network. AbMIL and CLAM are local attention networks because they score each image patch independently. Compensating the non-local interactions among patches should work better because, imitating the pathologist practice, we should put all the context of a slide to make decisions. As such, DSMIL uses a non-local stream to implement this intuition. TransMIL makes use of the transformer encoder to model these interactions. For our method, graph convolution is adopted to model the slide context, and we consider the multi-scale information at different magnification levels. The results indicated that our model performed better in predicting TMB status from WSIs. We attribute the consistent improvements of our method on all cancer types to our deliciated designs including feature embeddings, multi-level feature representation, and context-aware graph convolution.

Furthermore, we identified the improvements in identifying TMB status by MC-TMB over the single-modality tumor models. The critical challenge for predicting TMB status consists of the balance between sensitivity and specificity consisting of WES. The improvements in prediction accuracy in the multiple cancer model might profit from the increased training dataset and the cross talk of image features from multiple solid tumors. Our study provided an important hint for deep image learning based on the connectivity among various tumor types.

Deep learning approaches to a single dataset are prone to overfit and should be validated in external populations before clinical deployment. However, the generalizability of the prediction model is limited in previous studies. In this study, our MC-TMB algorithm also exhibited good generalization on external validation cohort, which might attribute to (1) effective feature representation with self-supervised learning performed on TCGA was much more robust than the ubiquitous transfer learning on ImageNet; (2) feature aggregation model using graph neural network and attention pooling to capture instance-level histology features and topological structures in the tumor microenvironment; (3) multi-scale prediction ensemble to take advantage of more complementary information at different resolutions.

Spatial heterogeneity of TMB within tumors was also identified through our MC-TMB. We used the proposed model pipeline to predict the TMB status of slides from 7 cancer types. The most attended regions recognized by MC-TMB were regarded to be associated with TMB-high (TMB-H). Areas with the red color of the heatmap represented the regions with predicted TMB-H status. The patch-level visualization by our model presented the spatial distributions of diverse mutational burden, which might potentially open avenues for exploring for genotype-spatial heterogeneity relationships.

Despite the preliminary application of WSI for predicting TMB in certain tumors, it is a pity that we had not yet applied deep learning to directly predict ICI outcomes. However, tumor immune microenvironment associated with MC-TMB (+) was identified in our study, which was relevant to microsatellite instability, lymphocyte infiltration, higher infiltration levels of CD8[+] T cells, and immune phenotypes associated with the response to cancer immunotherapy. Prediction of indirect outcome indicator might cause bias in primal intentions. Even though there are still challenges for the prediction of ICI reactiveness, our study had preliminarily confirmed the great potential of deep learning for the selection of candidates who might respond to ICI treatment based on histopathological images.

Limitations could also be found in this study. Firstly, although TMB is a biomarker for predicting immune therapy response, the application potential of MC-TMB in clinical practice still needs to be systematically evaluated and validated. Secondly, more external data sets and private patient cohorts are also wanted for further validation. Thirdly, there are only five types of tumors in the independent verification set, which might cause bias in the generalization performance of this model.

## CONCLUSION

Collectively, our study served as a proof-of-concept for developing multiscale, weakly-supervised deep learning framework for TMB status prediction and paved the way for prospective clinical trials to further assess the efficacy of deep learning-based TMB. Our study also identified the spatial heterogeneity of TMB within each tumor and enable image-based screening for molecular biomarkers with spatial variation and potential exploring for genotype-spatial heterogeneity relationships to facilitate personalized treatment decision-making.

## METHODS

### Study design and patient cohorts

In this retrospective study, we developed a weakly-supervised deep learning framework for developing and verifying MC-TMB from H&E-stained WSIs. The patient cohort was recruited from TCGA (https://portal.gdc.cancer.gov/) for this study. In addition, an independent cohort (CPTAC cohort) was collected from the Cancer Imaging Archive (https://www.cancerimagingarchive.net/). All recruited patients shall meet the following selection criteria: (i) pathologically diagnosed as single cancer without other types of malignant tumors; (ii) with corresponding clinical and pathological information; (iii) with access to diagnostic H&E-Stained WSIs; (iv) with retrievable TMB or total mutation count data (Fig. S3). The work has been reported in line with the STROCSS criteria.

**Patient-level classification**

TMB of the patient from TCGA was acquired from the cBioPortal for cancer genomics [28, 29], which was defined as nonsynonymous coding mutations per Megabase (Mb). Since currently there was no consensus on the determined value of TMB-H and TMB-low (TMB-L), the cut-off value for patient stratification was defined as 10 mutations per Mb (mut/Mb) according to previous reports for this study [3, 30]. Patients with nonsynonymous TMB of more than 10 mut/Mb were grouped into TMB-H. The independent validation cohort contained only total mutation count data. It would cause bias by setting the cut-off value as 10 mutations. Therefore, we performed the correlation analysis in the TCGA cohort and found that the total mutation count was highly correlated to TMB (related coefficient: 0.98, p < 0.0001, Fig. S4A). When the cut-off value of TMB was set as 10 mut/Mb, the relative total mutation count value was found to be 279 for the same percentage of TMB-H (Fig. S4B). So, the cut-off value of the total mutation count of TMB-H and TMB-L was set as 279 for the independent validation. After excluding some tumor types with a very low percentage of TMB-H and small data volume, we selected 7 types of cancer with a proportional of TMB-H for analysis, including COAD, STAD, LUAD, lung squamous cell carcinoma (LUSC), BLCA, HNSC, and uterine corpus endometrial carcinoma (UCEC).

**Feature embedding**

The gigapixel WSIs were loaded into memory at a downscaled level and then automated segmented with Otsu's method to exclude non-tissue regions. After region selection, we cropped the WSIs into tiles of non-overlapping 256 × 256 pixels within the foreground regions at user-customed magnification levels, i.e., ×5, ×10, and ×20 in our scenarios. To preserve the informative representation of raw pixels, ResNet50 was pretrained on the large-scale TCGA dataset following the unsupervised contrastive learning paradigm to extract robust and universal features across different tumor types from H&E slides (Fig. 1) [31, 32]. Following patching, we used the pretrained ResNet50 to compute a relatively low-dimensional feature representation for each cropped patch. The use of extracted features, i.e. 2048-dimensional, made it tractable to train the deep learning models with all tiles in a slide (up to 10,000 patches or more in a slide) simultaneously on a graphics processing unit (GPU).

**Feature convolution and aggregation**

After feature embedding, the following models extracted information from features via graph convolution and attention-based aggregation. The feature embedding from pretrained CNN could describe the local morphology of patches. However, a single-stream path of 256 × 256 pixels did not capture adequate spatial context from the tissue micro-environments. Graph-based representations could effectively describe the tissue composition by incorporating topology and interactions among entities, i.e. patches from the same slide. To construct the global graph representation of a slide, we saved (x, y) coordinates of cropped patches $X_j$ in raw images to build the adjacency matrix $A_j$ for each slide via fast approximate nearest neighbors k-NN (k=8). A patient-level slide could be denoted as CNN features and graph features $(X_j, A_j)$. Graph convolution network (GCN) updated a node's feature vector by aggregating feature information from its neighbors in the graph. We adapt DeepGCN [33] to process the CNN features $X_j$ along with its corresponding graph adjacent matrix $A_j$ following the ordering: normalization, ReLU activation, GraphConv, and addition. Empirically, the last step addition is residual learning which helps train deep networks. After graph convolution, the raw feature embedding $X_j \in \mathbb{R}^{N \times 2048}$ is translated into graph representation $G_j \in \mathbb{R}^{N \times 512}$, supposing that the initial embedding dimension is 2048, the translated dimension was 512, and N was the number of patches in a slide. After graph convolution, the following feature aggregation module was built around the trainable and interpretable attention-based function [13] to aggregate slide-level representation into a single vector $V_j$ for classification.

**Multi-scale classifier**

We aggregated the slide representations and the corresponding class label to produce a single probability per patient per magnification. In this retrospective study, we investigated TMB prediction and supposed that the cancer type of each slide could be exploited as additional information, providing cancer-specific clues such as visual representation heterogeneity of tumors for deep learning model. The class token was encoded in a one-hot vector. Then the attention-weighted feature vector $V_j$ and the class feature $C_j$ were fused by concatenation; the class token features were expanded by feeding into a fully connected layer to match the dimension of image features. Then, the fused features were fed into a classifier. The outputs and the ground-truth labels were used to calculate the cross-entropy loss for a given magnification level. Operating at a higher resolution captures local cellular information but limited the field-of-view due to computational burden and limits the access to global tissue microenvironment information. In contrast, operating at a lower resolution hindered resolvability of cells and access to cellular properties. We aggregated patient-level predictions from different magnifications i.e., ×5, ×10, and ×20 with a multiscale classifier.

**Model training**

Self-ensemble model training strategy was used to stabilize the training process. Each patch of H&E-stained WSI carried quantitative (not categorical positive or negative) information to the TMBs, with tumor architecture being a continuum. Thus, the supervision between label (i.e., positive or negative mutation burden) and input (i.e., H&E-stained WSI) was considered to be weak and inexact, compared with other classification tasks in natural images with distinctive labels. To stabilize the training procedure, we saved an exponential moving average version of the online training model. The aim was to establish ensemble learning as a backbone to form a solid consensus of the self-ensemble predictions [34]. The original code used in this study had been made available on GitHub (https://github.com/jinxixiang/MC-TMB)

**Statistical analysis**

In this study, receiver operating characteristic curve (ROC) analysis with an AUC and 95% confidence interval (CI) was carried out to evaluate the accuracy of the prediction model. Comparison between different AUCs was performed through a nonparametric approach reported by DeLong et al [35]. Correlation analysis was performed using the Spearman method. Continuous variant between the two groups was compared through the Mann-Whitney U test, and P value less than 0.05 was defined as significant. Kaplan-Meier analysis and log-rank tests were also applied to assess the survival outcomes among subgroups with MC-TMB (+) and MC-TMB (-), measured as hazard ratios (HR) with 95% CI.

**Tumor immune microenvironment analysis**

The mutation status of immunotherapy-related genes through whole genome sequencing was acquired from the cBioPortal [29] and analyzed via the *maftools* package [36] in R. The expressions of chemokines, interleukins, interferons, and major immune checkpoints were also retrieved from the cBioPortal. We carried out gene set enrichment analysis to enrich hallmark pathways relevant to MC-TMB (+) [37]. The abundances of immune cells and immune subtypes for each sample were estimated through CIBERSORT [38] and the immune landscape of cancer [19], respectively.

**AUTHOR CONTRIBUTIONS**

Conceptualization, J.Z., X.H., and J.H.Z.; methodology, J.X., X.W., and S.C.; investigation, S.Y., and J.H; statistical analyses, J.X., X.W., and S.C.; resources, W.Y.; writing – original draft, J.X., X.W., and S.C; writing – review & editing, J.X., X.W., S.C, J.Z., X.H., and J.H.Z.; funding acquisition, J.H.Z.; supervision, J.Z., X.H., and J.H.Z. X.H., and J.H.Z. had unrestricted access to all data. All authors read and approved the final article and take responsibility for its content.

**ACKNOWLEDGMENTS**

**CONFLICT OF INTEREST STATEMENT**

# Table

**Table 1**

Comparation of MC-TMB with other methods in the independent patient cohort.

|  | Aera under the curve | 95% confidence interval | p value |
|---|---|---|---|
| **MC-TMB** | 0.723 | 0.683-0.761 |  |
| **AbMIL** | 0.633 | 0.590-0.674 | < 0.0001 |
| **CLAM** | 0.651 | 0.609-0.692 | < 0.0001 |
| **DSMIL** | 0.670 | 0.628-0.710 | 0.0046 |
| **TransMIL** | 0.653 | 0.610-0.693 | 0.0001 |

**FIGURE LEGENDS**

**FIGURE 1** The network architecture of this study. (A) Feature embedding through unsupervised contrastive learning paradigm at multiple magnification level (×5, ×10, and ×20). (B) Graph convolution network updated a node's feature vector through aggregating feature information from its neighbors in the graph. (C) Feature convolution and aggregation for the multi-scale classification and further analyses. GCN, graph convolution network; CTF, Class token feature.

**FIGURE 2** The overall performance of the MC-TMB. (A) AUCs were evaluated in five-fold cross-validation in the internal validation cohort. (B) AUCs were evaluated in the independent validation cohort. AUC, area under curve; TMB, tumor mutational burden.

**FIGURE 3** Model performance at various magnifications. Area under curve was evaluated in five-fold cross-validation on ×5, ×10, ×20, and multiple magnifications, respectively.

**FIGURE 4** Analysis of attention heatmaps from MC-TMB. WSI, whole slide image; ROI, region of interest; TMB, tumor mutational burden; TMB-H, TMB-high; TMB-L, TMB-low.

**FIGURE 5** Comparison of multiple-cancer model and single-cancer model. (A) Multiple-cancer model outperformed single cancer model in predicting TMB status. (B) Comparison of the two models at various tumor types. (C) Investigate the advantage of multiple-cancer model through a detailed analysis of failure cases of single-cancer model. TMB, tumor mutational burden; WSI, whole slide image; ROI, region of interest.

**FIGURE 6** Molecular phenotypes associate with MC-TMB. (A) Distinct tumor mutation spectrums between MC-TMB (+) and MC-TMB (-). (B) MC-TMB (+) was associated with higher mutation frequency of other immunotherapy-related genes. (C) Performance of MC-TMB (+) in predicting the status of MSI and genes mutation. (D, E) Higher percentage of C2 immunophenotyping was found in MC-TMB (+).

**FIGURE 7** Tumor immune microenvironment associate with MC-TMB. (A) Different TMB and MSI-sensor score between MC-TMB (+) and MC-TMB (-). (B) MC-TMB (+) was significantly associated with higher infiltration levels of CD8$^+$ T cell. (C) Different lymphocyte infiltration signature and expression levels of PD-1/PD-L1 between MC-TMB (+) and MC-TMB (-). (D) Higher expression levels of chemokines were significantly associated with MC-TMB (+). (E) Kaplan-Meier survival analysis stratified by predicted TMB status. (F) Gene set enrichment analysis for the hallmark pathways associated with MC-TMB (+).